%% file: ms.tex
\documentclass{article}
\usepackage{spconf,amsmath,graphicx}
\usepackage[utf8]{inputenc}
\usepackage{pgfplots}
\usepackage{booktabs}

\title{MotifNet: a motif-based Graph Convolutional Network\\ for directed graphs}
\name{Federico Monti$^1$, Karl Otness$^2$, Michael M. Bronstein$^{1,2,3}$}
\address{$^1$University of Lugano\,\,\,\,\,   $^2$Harvard University\,\,\,\,\,   $^3$Tel Aviv University}
\usepackage{amssymb}
\newlength\fheight
\newlength\fwidth

\begin{document}
%
\maketitle
\begin{abstract}
Deep learning on graphs and in particular, graph convolutional neural networks, have recently attracted significant attention in the machine learning community. Many of such techniques explore the analogy between the graph Laplacian eigenvectors and the classical Fourier basis, allowing to formulate the convolution as a multiplication in the spectral domain. One of the key drawback of spectral CNNs is their explicit assumption of an undirected graph, leading to a symmetric Laplacian matrix with orthogonal eigendecomposition. 
In this work we propose MotifNet, a graph CNN capable of dealing with directed graphs by exploiting local graph motifs. We present experimental evidence showing the advantage of our approach on real data. 
\end{abstract}
\section{Introduction}
\label{sec:intro}

\input{introduction}

\section{Background}
\label{sec:background}

\input{background}

\begin{figure*}[t]
\centering
\setlength\fheight{4cm}
\setlength\fwidth{9.12cm}
\input{attentions_motifs_all_motifs_layer0_v2}	
\input{attentions_motifs_all_motifs_layer1_v2}	
\vspace*{-6mm} 
\caption{Attention scores $\alpha$ obtained with a MotifNet of order $p=1$ from the 1st and 2nd graph convolutional layers. Dark colors represent high probabilities, bright colors low ones. Only 7 of the 15 possible different motifs (considering the undirected adjacency matrix and the directed ones) appear as relevant for classifying the vertices of directed CORA. \vspace{-3mm}
\label{fig:architectureX}
}
\end{figure*}	

\section{MotifNet}
\label{sec:motifnet}

\input{motifnet_sec}

\vspace{-2.5mm}
\begin{figure}[!ht]
\includegraphics[width=9cm]{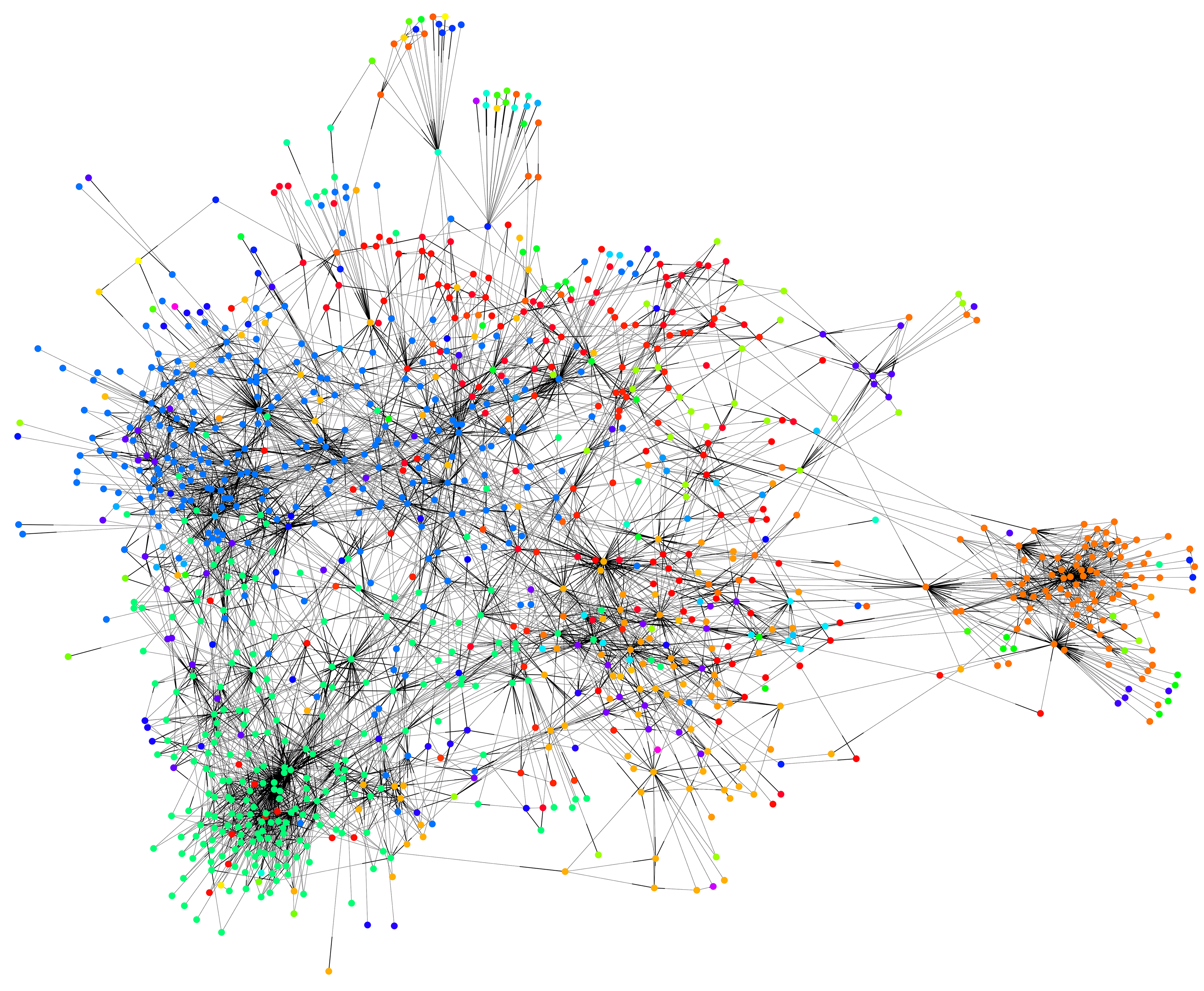}\vspace{-2.5mm}
\caption{Portion of the directed CORA dataset obtained extracting 1,000 vertices from the biggest weakly connected component. Different colors represent different classes.}\vspace{-4mm}
\end{figure}


\section{Experimental evaluation}
\label{sec:experiments}

\input{experiments}


\bibliographystyle{IEEEbib}
\bibliography{refs}

\end{document}

%% file: introduction.tex
Deep learning on graph-structured data has recently gained popularity in the machine learning community due to the increased interest in dealing with applications such as social network analysis and recommendation systems. 
One of the key challenges of generalizing successful deep neural network architectures such as convolutional neural networks (CNNs) to graphs is the lack of vector space structure and shift-invariance, resulting in the need to re-invent the basic building blocks of CNNs, including  convolutional filters and pooling.

Broadly speaking, we can distinguish between two classes of graph CNN formulations \cite{bronstein2017geometric}. 
Spatial approaches \cite{masci2015geodesic,boscaini2016learning,monti2016geometric} generalize the notion of `patch of pixels' by constructing a local system of weights on the graph. 
Spectral approaches \cite{bruna2013spectral,henaff2015deep,defferrard2016convolutional,kipf,monti2017geometric,levie2017cayleynets} use the analogy between the eigenfunctions of the graph Laplacian and the classical Fourier transform, and define a convolution-like operation in the spectral domain. 
So far, these methods were limited to undirected graphs, a restriction arising from the requirement to have a symmetric Laplacian matrix in order to obtain orthogonal eigendecomposition. At the same time, a wide variety of graph data, including citation networks, are directed, which limits the application of existing methods.

In this paper, we introduce MotifNet, a graph CNN for directed graphs. Our approach uses convolution-like anisotropic graph filters bases on local sub-graph structures (motifs) \cite{milo2002network,benson2016higher}. 
%
%
We use an attention mechanism, allowing MotifNet to generalize some standard graph CNN models without significantly increasing the model complexity. 
Experimental validation on real data shows superior performance compared to previous approaches.

%% file: background.tex
Let us be given a weighted undirected graph $\mathcal{G} = \{ \mathcal{V}, \mathcal{E}, \mathbf{W} \}$ with vertices $\mathcal{V} = \{1, \hdots, n\}$, edges $\mathcal{E} \subseteq \mathcal{V} \times \mathcal{V}$ s.t. $(i,j) \in \mathcal{E}$ iff $(j,i)\in \mathcal{E}$, and edge weights $w_{ij} \geq 0$ for $(i,j) \in \mathcal{E}$ and zero otherwise. 
The graph structure is represented by the $n\times n$ symmetric adjacency matrix $\mathbf{W} = (w_{ij})$.
We define the normalized graph Laplacian $\mathbf{\Delta} = \mathbf{I} - \mathbf{D}^{-1/2}\mathbf{W}\mathbf{D}^{-1/2}$, where $\mathbf{D} = \mathrm{diag}(\sum_{j \neq 1} w_{1j}, \hdots, \sum_{j \neq n} w_{nj})$ denotes the degree matrix.  
In the above setting, the Laplacian is a symmetric matrix and admits an eigendecomposition 
 $\mathbf{\Delta} = \mathbf{\Phi \Lambda \Phi}^\top$ with orthonormal eigenvectors $\boldsymbol{\Phi} = (\boldsymbol{\phi}_1^\top, \hdots, \boldsymbol{\phi}_n^\top)$ and non-negative eigenvalues $0 = \lambda_1 \leq \lambda_2 \leq  \hdots  \lambda_n$ arranged into a diagonal matrix $\boldsymbol{\Lambda} = \mathrm{diag}(\lambda_1, \hdots, \lambda_n)$.

We are interested in manipulating functions $f : \mathcal{V} \rightarrow \mathbb{R}$ defined on the vertices of the graph, which can be represented as vectors $\mathbf{f} \in \mathbb{R}^n$ and form a Hilbert space with the standard inner product $\langle \mathbf{f}, \mathbf{g} \rangle = \mathbf{f}^\top \mathbf{g}$. 
The eigenvectors of the Laplacian form an orthonormal basis in the aforementioned space of functions, allowing a Fourier decomposition of the form $\mathbf{f}  = \boldsymbol{\Phi} \boldsymbol{\Phi}^\top \mathbf{f}$,  where $\hat{\mathbf{f}} = \boldsymbol{\Phi}^\top \mathbf{f}$ is the {\em graph Fourier transform} of $\mathbf{f}$. The Laplacian eigenvectors thus play the role of the standard Fourier atoms and the corresponding eigenvalues that of frequencies. 
Finally, a convolution operation can be defined in the spectral domain by analogy to the Euclidean case as $\mathbf{f} \star \mathbf{g} =  \mathbf{\Phi} ( \hat{\mathbf{f}}  \cdot \hat{\mathbf{g}} ) = \mathbf{\Phi} (\mathbf{\Phi}^\top \mathbf{f}) \cdot (\mathbf{\Phi}^\top \mathbf{g}) $


Bruna et al. \cite{bruna2013spectral} exploited the above formulation for designing graph convolutional neural networks, in which a basic layer has the following form:
\begin{equation} 
\label{spectral_construction_eq}
\tilde{\mathbf{f}}_l =   \xi \left(  \sum_{l'=1}^{q'} \boldsymbol{\Phi} \hat{\mathbf{G}}_{ll'} \boldsymbol{\Phi}^\top \mathbf{f}_{l'} \right), \hspace{3mm} l = 1,\hdots, q,
\end{equation}
where $q', q$ denote the number of input and output channels, respectively, $\hat{\mathbf{G}}_{ll'} = \mathrm{diag}(\hat{g}_{ll',1}, \hdots, \hat{g}_{ll',n})$ is a diagonal matrix of spectral multipliers representing the filter, and $\xi$ is a nonlinearity (e.g. ReLU). 
Among the notable drawbacks of this architecture putting it at a clear disadvantage compared to classical Euclidean CNNs is high computational complexity ($\mathcal{O}(n^2)$ due to the cost of computing the forward and inverse graph Fourier transform, incurring dense $n\times n$ matrix multiplication), $\mathcal{O}(n)$ parameters per layer, and no guarantee of spatial localization of the filters. 
%
In order to cope with the two latter problems, Henaff et al. \cite{henaff2015deep} argued that filter localization is achieved by smoothness of its Fourier transform, and proposed parametrizing the filter as a smooth spectral transfer function. In particular, filters of the form $\hat{g}_k = \tau_{\boldsymbol{\theta}}(\lambda_k) = \sum_{j=1}^p \theta_j \beta_j(\lambda_k)$ were considered, where $\boldsymbol{\theta} = (\theta_1, \theta_2, \ldots, \theta_p)$ are the learnable filter parameters and $\beta_1(\lambda), \hdots, \beta_p(\lambda)$ are spline basis functions.


Defferrard et al. \cite{defferrard2016convolutional} considered the spectral CNN framework with polynomial filters represented in the Chebyshev basis (referred to as ChebNet), which can be efficiently computed by applying powers of the graph Laplacian 
\begin{equation} \label{eq:filt_cheby}
	\tilde{\mathbf{f}} = \mathbf{\Phi} \sum_{j=0}^{p} \theta_j T_j(\tilde{\boldsymbol{\Lambda}}) \mathbf{\Phi}^\top \mathbf{f} = \sum_{j=0}^{p} \theta_j T_j(\mathbf{\tilde{\Delta}})  \mathbf{f},
\end{equation}
and thus avoiding its eigendecomposition altogether. The computational complexity thus drops from $\mathcal{O}(n^2)$ to $\mathcal{O}(| \mathcal{E} |)$, and if the graph is sparsely connected, to $\mathcal{O}(n)$ 
%
%
%
%
(here $\tilde{\lambda}$ is a frequency rescaled in $[-1,1]$, $\tilde{\boldsymbol{\Delta}} = 2 \lambda_{n}^{-1}\boldsymbol{\Delta}  - \mathbf{I}$ is the rescaled Laplacian with eigenvalues $\tilde{\boldsymbol{\Lambda}} = 2 \lambda_{n}^{-1} \boldsymbol{\Lambda}  - \mathbf{I}$, and $T_j(\lambda) = 2\lambda T_{j-1}(\lambda) - T_{j-2}(\lambda)$ denotes the Chebyshev polynomial of degree $j$, with $T_1(\lambda) =\lambda$ and $T_0(\lambda) =1$).  

Kipf and Welling \cite{kipf} proposed a simplification of ChebNet (referred to as Graph Convolutional Network or GCN) by limiting the order of the polynomial to $p=1$ and using a re-normalization of the Laplacian to avoid numerical instability. 
%
%
Despite the efficiency of ChebNet \cite{defferrard2016convolutional} and GCN \cite{kipf}, both methods struggle when dealing with graphs containing clustered eigenvalues, a phenomenon typical in community graphs. Levie et al. \cite{levie2017cayleynets} used rational filter functions based on the Cayley transform, allowing to achieve better spectral resolution of the filters.



\section{DEALING WITH DIRECTED GRAPHS}

\begin{figure}[!ht]
\vspace{-1mm}
\centering 
\includegraphics[width=1\linewidth]{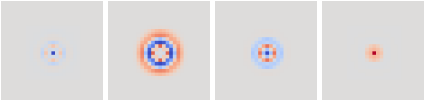}
\vspace{-6mm}
\caption{\label{fig:cheby}Examples of Chebyshev filters of degree $p=7$ on a regular grid. Note that the filters are isotropic due to rotational invariance of the Laplacian.}\vspace{-1mm}
\label{fig:MNIST-filters}
\end{figure}

One of the key drawbacks of the above spectral constructions is the explicit assumption of an undirected graph -- indeed, the existence of an orthonormal eigendecomposition of the Laplacian matrix crucially depends on the adjacency matrix $\mathbf{W}$ being symmetric, a property that is violated when the graph is directed. 
A further drawback is that the Laplacian operator is {\em isotropic}, i.e., has no preferred direction on the graph; consequently, the resulting spectral filters are rotationally symmetric when the underlying graph is a grid (see Figure \ref{fig:cheby}). While a construction of anisotropic Laplacians and thus oriented filters is possible on manifolds due to a locally-Euclidean structure \cite{boscaini2016learning}, it is more challenging on general graphs. 

\begin{figure}[!ht]
    \includegraphics[scale=0.6]{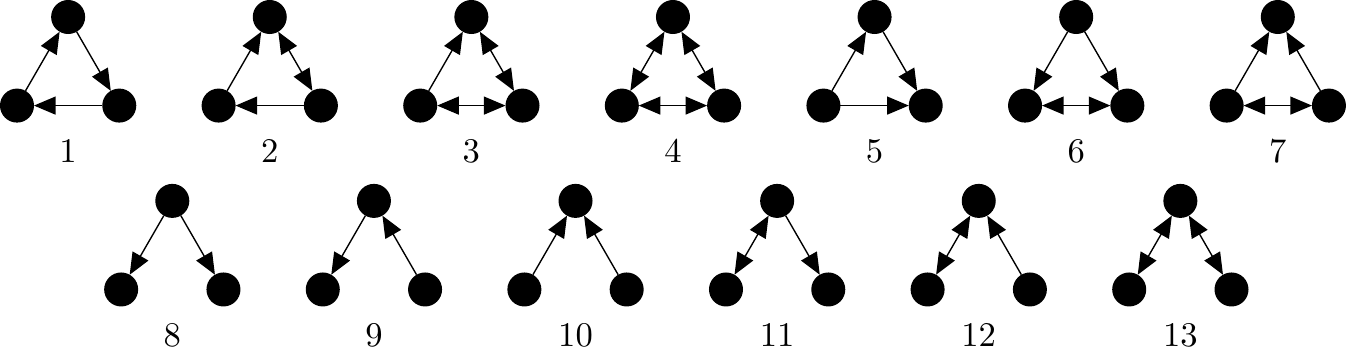}\vspace{-2.5mm}
\caption{\label{fig:motifs}Thirteen 3-vertex graph motifs used in this paper.}\vspace{-3mm}
\end{figure}

Benson et al. \cite{benson2016higher} proposed an elegant workaround these issues based on the analysis of small subgraphs called {\em motifs}. Let $\mathcal{G} = \{ \mathcal{V}, \mathcal{E}, \mathbf{W} \}$ be a weighted directed graph (in which case $\mathbf{W}^\top \neq \mathbf{W}$), and let $\mathcal{M}_1, \hdots, \mathcal{M}_K$ denote a collection of {\em graph motifs} (small directed graphs representing certain meaningful connectivity patterns; e.g., Figure \ref{fig:motifs} depicts thirteen 3-vertex motifs).  
For each edge $(i,j) \in \mathcal{E}$ of the directed graph $\mathcal{G}$ and each motif $\mathcal{M}_k$, let $u_{k,ij}$ denote the number of times the edge $(i,j)$ participates in $\mathcal{M}_k$ (note that an edge can participate in multiple motifs). Benson et al.  \cite{benson2016higher} define a new set of edge weights of the form $\tilde{w}_{k,ij} = u_{k,ij} w_{ij}$, which is now a {\em symmetric motif adjacency} matrix we denote by $\tilde{\mathbf{W}}_k$. 
The {\em motif Laplacian} $\tilde{\mathbf{\Delta}}_k = {\mathbf{I}} - \tilde{\mathbf{D}}_k^{-1/2}\tilde{\mathbf{W}}_k\tilde{\mathbf{D}}_k^{-1/2}$ associated with this adjacency acts {\em anisotropically} with a preferred direction along structures associated with the respective motif. 


%% file: attentions_motifs_all_motifs_layer0_v2.tex
\begin{tikzpicture}

\pgfplotsset{tick label style={font=\fontsize{5.5}{9}\selectfont}  
                    }

\begin{axis}[
xlabel={\small Motifs},
ylabel={\small Features},
xmin=-0.5, xmax=127.5,
ymin=-0.5, ymax=129.5,
width=\fwidth,
height=\fheight,
xtick={3,11,19,27,35,43,51,59,67,75,83,91,99,107,115,123},
ytick={0, 20, 40, 60, 80, 100, 120},
xticklabels={U,M$_1$,M$_2$,M$_3$,M$_4$,M$_5$,M$_6$,M$_7$,M$_8$,M$_9$,M$_{10}$,M$_{11}$,M$_{12}$,M$_{13}$,M$_{\mathrm{in}}$,M$_{\mathrm{out}}$},
tick align=outside,
tick pos=left,
x grid style={lightgray!92.02614379084967!black},
y grid style={lightgray!92.02614379084967!black},
point meta min=0,
point meta max=1,
colorbar style={ytick={0,0.2,0.4,0.6,0.8,1},yticklabels={0.0,0.2,0.4,0.6,0.8,1.0},ylabel={}},
ylabel style={yshift=-0.3cm}, 
]
\addplot graphics [includegraphics cmd=\pgfimage,xmin=-0.5, xmax=127.5, ymin=129.5, ymax=-0.5] {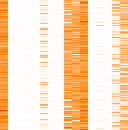};
\end{axis}

\end{tikzpicture}

%% file: attentions_motifs_all_motifs_layer1_v2.tex
\begin{tikzpicture}
\pgfplotsset{tick label style={font=\tiny}  
                    }
                    
\begin{axis}[
xlabel={\small Motifs},
ylabel={\small Features},
xmin=-0.5, xmax=127.5,
ymin=-0.5, ymax=129.5,
width=\fwidth,
height=\fheight,
xtick={3,11,19,27,35,43,51,59,67,75,83,91,99,107,115,123},
ytick={0, 20, 40, 60, 80, 100, 120},
xticklabels={U,M$_1$,M$_2$,M$_3$,M$_4$,M$_5$,M$_6$,M$_7$,M$_8$,M$_9$,M$_{10}$,M$_{11}$,M$_{12}$,M$_{13}$,M$_{\mathrm{in}}$,M$_{\mathrm{out}}$},
ytick={-20,0,20,40,60,80,100,120,140},
minor xtick={},
minor ytick={},
tick align=outside,
tick pos=left,
x grid style={lightgray!92.02614379084967!black},
y grid style={lightgray!92.02614379084967!black},
point meta min=0,
point meta max=1,
colorbar style={ytick={0,0.2,0.4,0.6,0.8,1},yticklabels={0.0,0.2,0.4,0.6,0.8,1.0},minor ytick={},ylabel={}},
ylabel style={yshift=-0.3cm}, 
]
\addplot graphics [includegraphics cmd=\pgfimage,xmin=-0.5, xmax=127.5, ymin=129.5, ymax=-0.5] {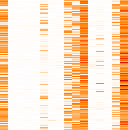};
\end{axis}

\end{tikzpicture}

%% file: motifnet_sec.tex
The key idea of this paper is using motif-induced adjacencies in the context of deep learning on graphs. We construct filters on the graph using {\em multivariate polynomial} filters of degree $p$ applied to the motif Laplacian matrices. 
Note that since the matrix product is generally non-commutative (i.e., $\tilde{\boldsymbol{\Delta}}_i \tilde{\boldsymbol{\Delta}}_j \neq \tilde{\boldsymbol{\Delta}}_j \tilde{\boldsymbol{\Delta}}_i$), we have $K^p$ products of the form $\tilde{\boldsymbol{\Delta}}_{k_1} \cdots  \tilde{\boldsymbol{\Delta}}_{k_p}$, where $k_l \in \{1, \hdots, K\}$. 
A general multivariate matrix polynomial has the form 
\begin{equation} 
\label{eq:mvpoly}
\mathbf{P}_{\boldsymbol{\Theta}}(\tilde{\boldsymbol{\Delta}}_1, \hdots, \tilde{\boldsymbol{\Delta}}_K) = \sum_{j=0}^{p} \,\, \sum_{k_1, \hdots, k_j \in \{1,\hdots, K \}}  \hspace{-6mm}\theta_{k_1, \hdots, k_j}  \tilde{\boldsymbol{\Delta}}_{k_1} \cdots  \tilde{\boldsymbol{\Delta}}_{k_j}, 
\end{equation}
where our convention is that for $j=0$ we have only one zero-degree term $\theta_0 \mathbf{I}$, and ${\boldsymbol{\Theta}}$ denotes the set of all the coefficients. 
Overall, a polynomial of the form~(\ref{eq:mvpoly}) has $\frac{1+K^{p+1}}{1-K}$ coefficients, which is impractically large even for a modest number of motifs $K$ or degree $p$.

We therefore study two possible simplifications of~(\ref{eq:mvpoly}). First, we consider only $K=2$ simple motifs corresponding to incoming and outgoing edges from a vertex. In this case, the polynomial becomes 
\begin{eqnarray} 
\label{eq:mvpoly1}
\mathbf{P}_{\boldsymbol{\Theta}} = \theta_0 \mathbf{I} + \theta_{1}\tilde{\boldsymbol{\Delta}}_1 + \theta_{2}\tilde{\boldsymbol{\Delta}}_2 + \theta_{11}\tilde{\boldsymbol{\Delta}}_1^2 + \hdots + \theta_{22}\tilde{\boldsymbol{\Delta}}_2^2 + \hdots 
\end{eqnarray}

Second, we consider a simplified version of multivariate polynomials~(\ref{eq:mvpoly}) defined recursively in the following manner,
\begin{eqnarray} 
\label{eq:mvpoly2}
\hspace{-3mm} \mathbf{P}_{\boldsymbol{\Theta}} (\tilde{\boldsymbol{\Delta}}_1, \hdots, \tilde{\boldsymbol{\Delta}}_K) \hspace{-1.5mm}&=&\hspace{-1.5mm} \sum_{j=0}^p \theta_j \mathbf{P}_j; \\
\hspace{-3mm}\mathbf{P}_{j}(\tilde{\boldsymbol{\Delta}}_1, \hdots, \tilde{\boldsymbol{\Delta}}_K) \hspace{-1.5mm}&=&\hspace{-1.5mm} \sum_{k=1}^K \alpha_{k,j} \tilde{\boldsymbol{\Delta}}_k \mathbf{P}_{j-1}, \hspace{1.5mm} j=1,\hdots, p \nonumber \\
\mathbf{P}_0 \hspace{-1.5mm}&=&\hspace{-1.5mm}  \mathbf{I}, \nonumber 
\end{eqnarray}
where $0 \leq \alpha_{i,j} \leq 1$ and $\boldsymbol{\Theta} = (\theta_0, \hdots, \theta_p, \alpha_{1,1}, \hdots, \alpha_{K,p})$ 
denotes the set of coefficients, $Kp+1$ in total.

{\em MotifNet} is a neural network architecture employing convolutional layers of the form 
\begin{equation} 
\label{eq:motifnet}
\tilde{\mathbf{f}}_l =  \xi \left( \sum_{ l'=1}^{q'}  \mathbf{P}_{ \boldsymbol{\Theta}_{ll'} } (\tilde{\boldsymbol{\Delta}}_1, \hdots, \tilde{\boldsymbol{\Delta}}_K)   \mathbf{f}_{l'} \right),\hspace{3mm} l = 1,\hdots, q,
\end{equation}
where $q', q$ denote the number of input and output channels, respectively, and $ \mathbf{P}_{ \boldsymbol{\Theta}_{ll'} }$ is the simplified multivariate matrix polynomial~(\ref{eq:mvpoly1}) or~(\ref{eq:mvpoly2}). 
ChebNet is obtained as a particular instance of MotifNet with a single Laplacian of an undirected graph, in which case a univariate matrix polynomial is used.

%% file: experiments.tex
We tested our approach on the directed CORA citation network \cite{bojchevski2017deep}. The vertices of the CORA graph represent 19,793 scientific papers, and directed edges of the form $(i,j)$ represent citation of paper $j$ in paper $i$. 
The content of each paper is represented by a vector of 8,710 numerical features (term frequency-inverse document frequency of various words that appear in the corpus), to which we applied PCA taking the first 130 components. The task is to classify the papers into one of the 70 different categories.


We considered the semi-supervised learning setting \cite{kipf} using 10\% of the available vertices for training, 10\% for validation and 10\% for testing. 
%
We compared the following graph CNN architectures: ChebNet (applied to an undirected version of CORA and to the directed adjacency matrices $\mathbf{W}$ and $\mathbf{W}^\top$) and two versions of MotifNet using simplified multivariate matrix polynomial~(\ref{eq:mvpoly1}) or~(\ref{eq:mvpoly2}), to which we refer as MotifNet-m and MotifNet-d, respectively. All the graph CNNs contained two convolutional layers and a final fully connected layer followed by softmax. 
MotifNet-d contained only in/outgoing edges (denoted by $\mathcal{M}_\mathrm{in}$ and $\mathcal{M}_\mathrm{out}$). 
For MotifNet-m, we considered 13 motifs formed by triplets of vertices (shown in Figure \ref{fig:motifs} and denoted $\mathcal{M}_1, \hdots, \mathcal{M}_{13}$). 
To reduce the computational complexity of our model, we selected a subset of motifs in the following way. First, we trained MotifNet with order $p=1$ and selected the model with minimum cross-entropy on the validation set. By analyzing the probabilities learned by our model,  we discovered that only several motifs (depicted in Figure \ref{fig:motifs}) turned out to be relevant. 
Then, we used this subset of motifs $\mathcal{M}_5, \mathcal{M}_8, \mathcal{M}_9$  in addition to the undirected graph obtained by replacing each directed edge with an undirected one (denoted by $\mathcal{U}$) and $\mathcal{M}_\mathrm{in}$ and $\mathcal{M}_\mathrm{out}$\footnote{$\mathcal{M}_{10}$ has been discarded in our final architecture because of the dense motif adjacency matrix it was presenting.}.

%

All models were trained on NVIDIA Titan X GPU. 
Dropout with keep probability of $0.5$ and weight decay with constant $\gamma = 10^{-3}$ were used as regularization.  Adam \cite{KingmaB14} optimization method was used to train the models with learning rate equal to $10^{-3}$.



\begin{figure}[!h]
\setlength\fheight{4cm}
\setlength\fwidth{8.5cm}
\vspace{-1mm}
\input{./plot_chebnet_motifnet.tikz}\vspace{-3.5mm}
\caption{\label{fig:res1}Classification accuracy on CORA obtained with ChebNet on undirected graph (blue) and MotifNet-m (orange). }\vspace{-2.5mm}
\end{figure}
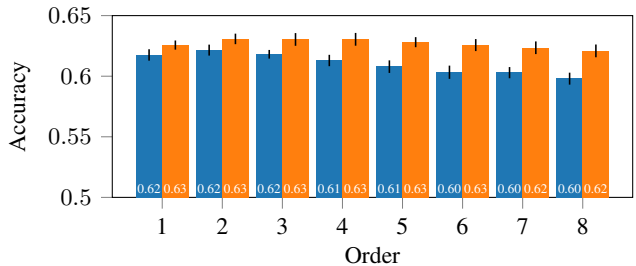

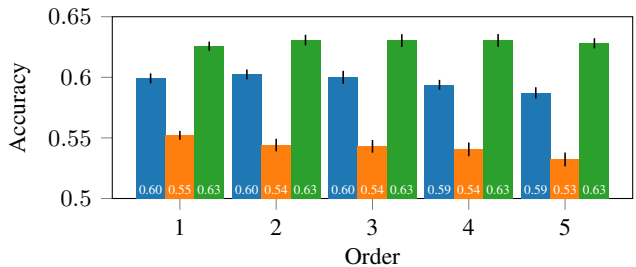
\begin{figure}[!h]
\setlength\fheight{4cm}
\setlength\fwidth{8.5cm}
\vspace{-1mm}
\input{./plot_dir_multivar_motifnet.tikz}\vspace{-3.5mm}
\caption{\label{fig:res2}Classification accuracy on CORA obtained with ChebNet applied with adjacency matrix $\mathbf{W}$ (blue) / $\mathbf{W}^\top$ (red), MotifNet-d (green) and MotifNet-m (orange).}\vspace{-2.5mm}
\end{figure}

The results we obtained are reported in Figures \ref{fig:res1}-\ref{fig:res2}. MotifNet-m consistently outperforms the baseline (ChebNet) for a variety of different polynomial orders, at the expense of only a tiny increase in the number of parameters (Table \ref{tab:params}).

\begin{table}[]
\centering
\caption{\label{tab:params}Number of parameters required by the considered GCNs. MotifNet requires just a handful of additional parameters for handling the considered adjacency matrices. }
\label{my-label}
\begin{tabular}{@{}ccccc@{}}
\toprule
Order  & ChebNet & MotifNet-m & MotifNet-d\\ \midrule
1 & 94K   & 95K & 128K  \\
2 & 128K  & 131K  & 263K   \\
3 & 162K  & 166K  & 534K   \\
4 & 196K  & 202K  & 1,074K  \\
5 & 229K   & 237K  & 2,156K  \\
6 & 263K   & 272K & 4,319K  \\
7 & 297K   & 308K  & 8,646K  \\
8 & 331K   & 343K  & 17,298K \\ \bottomrule
\end{tabular}\vspace{-2mm}
\end{table}

\vspace{-2mm}
\vspace{-3.2mm}

%% file: plot_chebnet_motifnet.tikz
\begin{tikzpicture}

\definecolor{color0}{rgb}{0.12156862745098,0.466666666666667,0.705882352941177}
\definecolor{color1}{rgb}{1,0.498039215686275,0.0549019607843137}

\begin{axis}[
xlabel={\small Order},
ylabel={\small Accuracy},
xmin=-3.325, xmax=31.325,
ymin=0.5, ymax=0.65,
width=\fwidth,
height=\fheight,
xtick={0,4,8,12,16,20,24,28},
xticklabels={1,2,3,4,5,6,7,8},
xticklabel style={font=\small},
yticklabel style={font=\small},
tick align=outside,
tick pos=left,
x grid style={white!69.01960784313725!black},
y grid style={white!69.01960784313725!black}
]
\draw[fill=color0,draw opacity=0] (axis cs:-1.75,0) rectangle (axis cs:0,0.61753714);
\draw[fill=color0,draw opacity=0] (axis cs:2.25,0) rectangle (axis cs:4,0.6215309);
\draw[fill=color0,draw opacity=0] (axis cs:6.25,0) rectangle (axis cs:8,0.6181004);
\draw[fill=color0,draw opacity=0] (axis cs:10.25,0) rectangle (axis cs:12,0.61295444);
\draw[fill=color0,draw opacity=0] (axis cs:14.25,0) rectangle (axis cs:16,0.607885);
\draw[fill=color0,draw opacity=0] (axis cs:18.25,0) rectangle (axis cs:20,0.603277);
\draw[fill=color0,draw opacity=0] (axis cs:22.25,0) rectangle (axis cs:24,0.602919);
\draw[fill=color0,draw opacity=0] (axis cs:26.25,0) rectangle (axis cs:28,0.597978);
\draw[fill=color1,draw opacity=0] (axis cs:0,0) rectangle (axis cs:1.75,0.625704);
\draw[fill=color1,draw opacity=0] (axis cs:4,0) rectangle (axis cs:5.75,0.630773);
\draw[fill=color1,draw opacity=0] (axis cs:8,0) rectangle (axis cs:9.75,0.630338);
\draw[fill=color1,draw opacity=0] (axis cs:12,0) rectangle (axis cs:13.75,0.63044);
\draw[fill=color1,draw opacity=0] (axis cs:16,0) rectangle (axis cs:17.75,0.628085);
\draw[fill=color1,draw opacity=0] (axis cs:20,0) rectangle (axis cs:21.75,0.625678);
\draw[fill=color1,draw opacity=0] (axis cs:24,0) rectangle (axis cs:25.75,0.623477);
\draw[fill=color1,draw opacity=0] (axis cs:28,0) rectangle (axis cs:29.75,0.620891);
\path [draw=black, semithick] (axis cs:-0.875,0.61282901)
--(axis cs:-0.875,0.62224527);

\path [draw=black, semithick] (axis cs:3.125,0.61700795)
--(axis cs:3.125,0.62605385);

\path [draw=black, semithick] (axis cs:7.125,0.61452615)
--(axis cs:7.125,0.62167465);

\path [draw=black, semithick] (axis cs:11.125,0.60826186)
--(axis cs:11.125,0.61764702);

\path [draw=black, semithick] (axis cs:15.125,0.602707)
--(axis cs:15.125,0.613063);

\path [draw=black, semithick] (axis cs:19.125,0.597823)
--(axis cs:19.125,0.608731);

\path [draw=black, semithick] (axis cs:23.125,0.598298)
--(axis cs:23.125,0.60754);

\path [draw=black, semithick] (axis cs:27.125,0.593037)
--(axis cs:27.125,0.602919);

\path [draw=black, semithick] (axis cs:0.875,0.6219309)
--(axis cs:0.875,0.6294771);

\path [draw=black, semithick] (axis cs:4.875,0.62642855)
--(axis cs:4.875,0.63511745);

\path [draw=black, semithick] (axis cs:8.875,0.62509253)
--(axis cs:8.875,0.63558347);

\path [draw=black, semithick] (axis cs:12.875,0.62516612)
--(axis cs:12.875,0.63571388);

\path [draw=black, semithick] (axis cs:16.875,0.62393698)
--(axis cs:16.875,0.63223302);

\path [draw=black, semithick] (axis cs:20.875,0.62073093)
--(axis cs:20.875,0.63062507);

\path [draw=black, semithick] (axis cs:24.875,0.61824706)
--(axis cs:24.875,0.62870694);

\path [draw=black, semithick] (axis cs:28.875,0.61562577)
--(axis cs:28.875,0.62615623);

\node at (axis cs:-0.875,0.5)[
  scale=0.5,
  anchor=south,
  text=white,
  rotate=0.0
]{ 0.62};
\node at (axis cs:3.125,0.5)[
  scale=0.5,
  anchor=south,
  text=white,
  rotate=0.0
]{ 0.62};
\node at (axis cs:7.125,0.5)[
  scale=0.5,
  anchor=south,
  text=white,
  rotate=0.0
]{ 0.62};
\node at (axis cs:11.125,0.5)[
  scale=0.5,
  anchor=south,
  text=white,
  rotate=0.0
]{ 0.61};
\node at (axis cs:15.125,0.5)[
  scale=0.5,
  anchor=south,
  text=white,
  rotate=0.0
]{ 0.61};
\node at (axis cs:19.125,0.5)[
  scale=0.5,
  anchor=south,
  text=white,
  rotate=0.0
]{ 0.60};
\node at (axis cs:23.125,0.5)[
  scale=0.5,
  anchor=south,
  text=white,
  rotate=0.0
]{ 0.60};
\node at (axis cs:27.125,0.5)[
  scale=0.5,
  anchor=south,
  text=white,
  rotate=0.0
]{ 0.60};
\node at (axis cs:0.875,0.5)[
  scale=0.5,
  anchor=south,
  text=white,
  rotate=0.0
]{ 0.63};
\node at (axis cs:4.875,0.5)[
  scale=0.5,
  anchor=south,
  text=white,
  rotate=0.0
]{ 0.63};
\node at (axis cs:8.875,0.5)[
  scale=0.5,
  anchor=south,
  text=white,
  rotate=0.0
]{ 0.63};
\node at (axis cs:12.875,0.5)[
  scale=0.5,
  anchor=south,
  text=white,
  rotate=0.0
]{ 0.63};
\node at (axis cs:16.875,0.5)[
  scale=0.5,
  anchor=south,
  text=white,
  rotate=0.0
]{ 0.63};
\node at (axis cs:20.875,0.5)[
  scale=0.5,
  anchor=south,
  text=white,
  rotate=0.0
]{ 0.63};
\node at (axis cs:24.875,0.5)[
  scale=0.5,
  anchor=south,
  text=white,
  rotate=0.0
]{ 0.62};
\node at (axis cs:28.875,0.5)[
  scale=0.5,
  anchor=south,
  text=white,
  rotate=0.0
]{ 0.62};
\end{axis}

\end{tikzpicture}

%% file: plot_dir_multivar_motifnet.tikz
\begin{tikzpicture}

\definecolor{color0}{rgb}{0.12156862745098,0.466666666666667,0.705882352941177}
\definecolor{color1}{rgb}{1,0.498039215686275,0.0549019607843137}
\definecolor{color2}{rgb}{0.172549019607843,0.627450980392157,0.172549019607843}

\begin{axis}[
xlabel={\small Order},
ylabel={\small Accuracy},
xmin=-4.0375, xmax=27.0375,
ymin=0.5, ymax=0.65,
width=\fwidth,
height=\fheight,
xtick={0,5.75,11.5,17.25,23},
xticklabels={1,2,3,4,5},
xticklabel style={font=\small},
yticklabel style={font=\small},
tick align=outside,
tick pos=left,
x grid style={white!69.01960784313725!black},
y grid style={white!69.01960784313725!black}
]
\draw[fill=color0,draw opacity=0] (axis cs:-2.625,0) rectangle (axis cs:-0.875,0.599104);
\draw[fill=color0,draw opacity=0] (axis cs:3.125,0) rectangle (axis cs:4.875,0.602381);
\draw[fill=color0,draw opacity=0] (axis cs:8.875,0) rectangle (axis cs:10.625,0.600026);
\draw[fill=color0,draw opacity=0] (axis cs:14.625,0) rectangle (axis cs:16.375,0.59383);
\draw[fill=color0,draw opacity=0] (axis cs:20.375,0) rectangle (axis cs:22.125,0.587225);
\draw[fill=color1,draw opacity=0] (axis cs:-0.875,0) rectangle (axis cs:0.875,0.552099);
\draw[fill=color1,draw opacity=0] (axis cs:4.875,0) rectangle (axis cs:6.625,0.544112);
\draw[fill=color1,draw opacity=0] (axis cs:10.625,0) rectangle (axis cs:12.375,0.543036);
\draw[fill=color1,draw opacity=0] (axis cs:16.375,0) rectangle (axis cs:18.125,0.540553);
\draw[fill=color1,draw opacity=0] (axis cs:22.125,0) rectangle (axis cs:23.875,0.532232);
\draw[fill=color2,draw opacity=0] (axis cs:0.875,0) rectangle (axis cs:2.625,0.625704);
\draw[fill=color2,draw opacity=0] (axis cs:6.625,0) rectangle (axis cs:8.375,0.630773);
\draw[fill=color2,draw opacity=0] (axis cs:12.375,0) rectangle (axis cs:14.125,0.630338);
\draw[fill=color2,draw opacity=0] (axis cs:18.125,0) rectangle (axis cs:19.875,0.63044);
\draw[fill=color2,draw opacity=0] (axis cs:23.875,0) rectangle (axis cs:25.625,0.628085);
\path [draw=black, semithick] (axis cs:-1.75,0.59499)
--(axis cs:-1.75,0.603218);

\path [draw=black, semithick] (axis cs:4,0.598319)
--(axis cs:4,0.606443);

\path [draw=black, semithick] (axis cs:9.75,0.594687)
--(axis cs:9.75,0.605365);

\path [draw=black, semithick] (axis cs:15.5,0.58975)
--(axis cs:15.5,0.59791);

\path [draw=black, semithick] (axis cs:21.25,0.582633)
--(axis cs:21.25,0.591817);

\path [draw=black, semithick] (axis cs:0,0.548431)
--(axis cs:0,0.555767);

\path [draw=black, semithick] (axis cs:5.75,0.538939)
--(axis cs:5.75,0.549285);

\path [draw=black, semithick] (axis cs:11.5,0.537836)
--(axis cs:11.5,0.548236);

\path [draw=black, semithick] (axis cs:17.25,0.534907)
--(axis cs:17.25,0.546199);

\path [draw=black, semithick] (axis cs:23,0.526531)
--(axis cs:23,0.537933);

\path [draw=black, semithick] (axis cs:1.75,0.6219309)
--(axis cs:1.75,0.6294771);

\path [draw=black, semithick] (axis cs:7.5,0.62642855)
--(axis cs:7.5,0.63511745);

\path [draw=black, semithick] (axis cs:13.25,0.62509253)
--(axis cs:13.25,0.63558347);

\path [draw=black, semithick] (axis cs:19,0.62516612)
--(axis cs:19,0.63571388);

\path [draw=black, semithick] (axis cs:24.75,0.62393698)
--(axis cs:24.75,0.63223302);

\node at (axis cs:-1.75,0.5)[
  scale=0.5,
  anchor=south,
  text=white,
  rotate=0.0
]{ 0.60};
\node at (axis cs:4,0.5)[
  scale=0.5,
  anchor=south,
  text=white,
  rotate=0.0
]{ 0.60};
\node at (axis cs:9.75,0.5)[
  scale=0.5,
  anchor=south,
  text=white,
  rotate=0.0
]{ 0.60};
\node at (axis cs:15.5,0.5)[
  scale=0.5,
  anchor=south,
  text=white,
  rotate=0.0
]{ 0.59};
\node at (axis cs:21.25,0.5)[
  scale=0.5,
  anchor=south,
  text=white,
  rotate=0.0
]{ 0.59};
\node at (axis cs:0,0.5)[
  scale=0.5,
  anchor=south,
  text=white,
  rotate=0.0
]{ 0.55};
\node at (axis cs:5.75,0.5)[
  scale=0.5,
  anchor=south,
  text=white,
  rotate=0.0
]{ 0.54};
\node at (axis cs:11.5,0.5)[
  scale=0.5,
  anchor=south,
  text=white,
  rotate=0.0
]{ 0.54};
\node at (axis cs:17.25,0.5)[
  scale=0.5,
  anchor=south,
  text=white,
  rotate=0.0
]{ 0.54};
\node at (axis cs:23,0.5)[
  scale=0.5,
  anchor=south,
  text=white,
  rotate=0.0
]{ 0.53};
\node at (axis cs:1.75,0.5)[
  scale=0.5,
  anchor=south,
  text=white,
  rotate=0.0
]{ 0.63};
\node at (axis cs:7.5,0.5)[
  scale=0.5,
  anchor=south,
  text=white,
  rotate=0.0
]{ 0.63};
\node at (axis cs:13.25,0.5)[
  scale=0.5,
  anchor=south,
  text=white,
  rotate=0.0
]{ 0.63};
\node at (axis cs:19,0.5)[
  scale=0.5,
  anchor=south,
  text=white,
  rotate=0.0
]{ 0.63};
\node at (axis cs:24.75,0.5)[
  scale=0.5,
  anchor=south,
  text=white,
  rotate=0.0
]{ 0.63};
\end{axis}

\end{tikzpicture}

%% file: ms.bbl
\begin{thebibliography}{10}

\bibitem{bronstein2017geometric}
M.~M. Bronstein, J.~Bruna, Y.~LeCun, A.~Szlam, and P.~Vandergheynst,
\newblock ``Geometric deep learning: going beyond euclidean data,''
\newblock {\em IEEE Signal Processing Magazine}, vol. 34, no. 4, pp. 18--42,
  2017.

\bibitem{masci2015geodesic}
J.~Masci, D.~Boscaini, M.~Bronstein, and P.~Vandergheynst,
\newblock ``Geodesic convolutional neural networks on riemannian manifolds,''
\newblock in {\em Proc. 3dRR}, 2015.

\bibitem{boscaini2016learning}
D.~Boscaini, J.~Masci, E.~Rodol{\`a}, and M.~M. Bronstein,
\newblock ``Learning shape correspondence with anisotropic convolutional neural
  networks,''
\newblock in {\em Proc. NIPS}, 2016.

\bibitem{monti2016geometric}
F.~Monti, D.~Boscaini, J.~Masci, E.~Rodol{\`a}, J.~Svoboda, and M.~M.
  Bronstein,
\newblock ``Geometric deep learning on graphs and manifolds using mixture model
  cnns,''
\newblock in {\em Proc. CVPR}, 2017.

\bibitem{bruna2013spectral}
J.~Bruna, W.~Zaremba, A.~Szlam, and Y.~LeCun,
\newblock ``Spectral networks and locally connected networks on graphs,''
\newblock {\em arXiv:1312.6203}, 2013.

\bibitem{henaff2015deep}
M.~Henaff, J.~Bruna, and Y.~LeCun,
\newblock ``Deep convolutional networks on graph-structured data,''
\newblock {\em arXiv:1506.05163}, 2015.

\bibitem{defferrard2016convolutional}
M.~Defferrard, X.~Bresson, and P.~Vandergheynst,
\newblock ``Convolutional neural networks on graphs with fast localized
  spectral filtering,''
\newblock in {\em Proc. NIPS}, 2016.

\bibitem{kipf}
T.~N. Kipf and M.~Welling,
\newblock ``Semi-supervised classification with graph convolutional networks,''
\newblock 2017.

\bibitem{monti2017geometric}
F.~Monti, M.~M. Bronstein, and X.~Bresson,
\newblock ``Geometric matrix completion with recurrent multi-graph neural
  networks,''
\newblock in {\em Proc. NIPS}, 2017.

\bibitem{levie2017cayleynets}
R.~Levie, F.~Monti, X.~Bresson, and M.~M Bronstein,
\newblock ``Cayleynets: Graph convolutional neural networks with complex
  rational spectral filters,''
\newblock {\em arXiv:1705.07664}, 2017.

\bibitem{milo2002network}
R.~Milo, S.~Shen-Orr, S.~Itzkovitz, N.~Kashtan, D.~Chklovskii, and U.~Alon,
\newblock ``Network motifs: simple building blocks of complex networks,''
\newblock {\em Science}, vol. 298, no. 5594, pp. 824--827, 2002.

\bibitem{benson2016higher}
A.~R. Benson, D.~F. Gleich, and J.~Leskovec,
\newblock ``Higher-order organization of complex networks,''
\newblock {\em Science}, vol. 353, no. 6295, pp. 163--166, 2016.

\bibitem{bojchevski2017deep}
Aleksandar Bojchevski and Stephan G{\"u}nnemann,
\newblock ``Deep gaussian embedding of attributed graphs: Unsupervised
  inductive learning via ranking,''
\newblock {\em arXiv:1707.03815}, 2017.

\bibitem{KingmaB14}
D.~P. Kingma and J.~Ba,
\newblock ``Adam: A method for stochastic optimization,''
\newblock in {\em Proc. ICLR}, 2015.

\end{thebibliography}
